\documentclass{article}
\usepackage{arxiv}
\usepackage[utf8]{inputenc} 
\usepackage[T1]{fontenc}    
\usepackage{hyperref}       
\usepackage{url}            
\usepackage{booktabs}       
\usepackage{amsfonts}       
\usepackage{amsmath}
\usepackage{amssymb}
\usepackage{nicefrac}       
\usepackage{microtype}      
\usepackage{lipsum}		    
\usepackage{graphicx}
\usepackage{natbib}
\usepackage{doi}
\usepackage{algorithm}
\usepackage{algpseudocode} 

\title{On efficient computation in active inference}

\date{\today}

\author{{Aswin Paul\textsuperscript{1, 2, 3}, Noor Sajid\textsuperscript{4}, Lancelot Da Costa\textsuperscript{4, 5, 6} and Adeel Razi\textsuperscript{1, 4, 7}} \\
    \\
    \textsuperscript{1} Turner Institute for Brain and Mental Health, Monash University, Clayton 3800, Australia \\
    \textsuperscript{2} IITB-Monash Research Academy, Mumbai, India \\
    \textsuperscript{3} Department of Electrical Engineering, IIT Bombay, Mumbai, India \\
    \textsuperscript{4} Wellcome Centre for Human Neuroimaging, University College London, WC1N 3AR London, United Kingdom \\
        \textsuperscript{5} Department of Mathematics, Imperial College London, London, SW7 2AZ, UK \\
    \textsuperscript{6} VERSES Research Lab, Los Angeles, CA 90016, USA \\
    \textsuperscript{7} CIFAR Azrieli Global Scholars Program, CIFAR, Toronto, Canada \\
}

\renewcommand{\shorttitle}{On efficient computation in active inference}
\algnewcommand{\algorithmicand}{\textbf{ and }}
\algnewcommand{\algorithmicor}{\textbf{ or }}
\algnewcommand{\OR}{\algorithmicor}
\algnewcommand{\AND}{\algorithmicand}
\algnewcommand{\var}{\texttt}

\begin{document}
\maketitle

\begin{abstract}
Despite being recognized as neurobiologically plausible, active inference faces difficulties when employed to simulate intelligent behaviour in complex environments due to its computational cost and the difficulty of specifying an appropriate target distribution for the agent. This paper introduces two solutions that work in concert to address these limitations. First, we present a novel planning algorithm for finite temporal horizons with drastically lower computational complexity. Second, inspired by Z-learning from control theory literature, we simplify the process of setting an appropriate target distribution for new and existing active inference planning schemes. 
Our first approach leverages the dynamic programming algorithm, known for its computational efficiency, to minimize the cost function used in planning through the Bellman-optimality principle.
Accordingly, our algorithm recursively assesses the expected free energy of actions in the reverse temporal order. This improves computational efficiency by orders of magnitude and allows precise model learning and planning, even under uncertain conditions. Our method simplifies the planning process and shows meaningful behaviour even when specifying only the agent's final goal state. 
The proposed solutions make defining a target distribution from a goal state straightforward compared to the more complicated task of defining a temporally informed target distribution. The effectiveness of these methods is tested and demonstrated through simulations in standard grid-world tasks. These advances create new opportunities for various applications. 
\end{abstract}




\section{Introduction}

How should an organism perceive, learn, and act to ensure survival when born into a new world? How do `agents' eventually learn to exhibit sentient behaviour in nature, such as hunting and navigation?

A prominent framework that approaches these questions is stochastic optimal control (SOC), which determines the best possible set of decisions---given a specific criterion---at any given time and in the face of uncertainty. The fundamental problem that SOC addresses can be defined as follows: When born at time $t=1$ and ahead, an `agent' receives observations from its surrounding `environment'. This `agent' not only passively receives observations but also is capable of responding with `actions'. Additionally, it may receive information or has inbuilt reward systems that quantify its chance of survival and progress. So, this process may be summarised as a stream of data from the agent's perspective: $(o_{1}; ~a_{1}), (o_{2}, ~r_{2}; ~a_2),..., (o_{t}, ~r_{t}) $. Here, $o_{t}$ stands for the observation at time $t$, $a_{t}$ stands for the agent's action at time $t$, and $r_{t}$ stands for the `reward' at time $t$ from the external environment or agent's inbuilt reward structure. In this setting, the primary goal of an agent is to

\begin{equation}\label{Eq:stochasticCP}
    \text{Maximise:~} \text{Score} = \sum_{1}^t r_{t}. \footnote{Reward scores the desirability for a particular outcome or state; akin to some cost function. Briefly, it can be explicitly defined by the 'external' environment (extrinsic reward) or internally by the agent itself (intrinsic reward).}
\end{equation}

Eq.\ref{Eq:stochasticCP} is an optimisation problem, and due to its general structure, it has a vast scope in various disciplines across the sciences. Several fields of research grew around this idea in the past decades, like reinforcement learning (RL) \citep{Sutton2018}, control theory \citep{Todorov2006, Todorov2009}, game theory \citep{fudenberg1991game,lu2020online}, and economics \citep{Mookherjee1984,vonneumannTheoryGamesEconomic1944}. But in fact, formulating decision-making as utility maximisation originated much earlier in the ethical theory of utilitarianism in 18th-century philosophy \citep{benthamIntroductionPrinciplesMorals1781,millUtilitarianism1870}, and was later applied by Pavlov in the early 20th century to account for animal conditioning \citep{pavlov1927ConditionedReflexesInvestigation2010}. Many current engineering methods, such as Q-learning \citep{watkins1992q}, build upon the Bellman-optimality principle to learn proper observation-action mappings that maximise cumulative reward. Model-based methods in RL, like Dyna-Q \cite{peng1993efficient}, employ an internal model of the `environment' to accelerate this planning process \cite{Sutton2018}. Similarly, efficient methods, e.g., which linearly scales with the problem dimensions, emerged in classical control theory to compute optimal actions in similar settings \citep{Todorov2006, Todorov2009}. 

Another critical and complementary research direction is studying systems showing `general intelligence', which abounds in nature. Indeed, we see a spectrum of behaviour in the natural world that may or may not be accountable by the rather narrow goal of optimising cumulative reward. By learning more about how the brain produces sentient behaviour, we can hope to accelerate the generation of artificial general intelligence \citep{goertzel2014artificial,gershmanWhatHaveWe2023}. This outlook motivates us to look into the neural and cognitive sciences, where an integral theory is the free energy principle (FEP), which brings together Helmholtz's early observations of perception with more recent ideas from statistical physics and machine learning \citep{feynmanStatisticalMechanicsSet1998,dayanHelmholtzMachine1995} to attempt a mathematical description of brain function and behaviour in terms of inference that has the potential of unifying many previous theories on the subject, including but not limited to cumulative reward maximisation \citep{fristonFreeenergyPrincipleUnified2010,fristonFreeEnergyPrinciple2022,dacostaRewardMaximizationDiscrete2023}.

In the last decade, the FEP has been applied to model and generate biological-like behaviour under the banner of \textit{active inference} \citep{DaCosta2020}. Active inference has since percolated into many adjacent fields owing to its ambitious scope as a general modelling framework for behaviour \citep{Pezzato_2023, Oliver_2022,Deane2020,Sergio2020,fountasDeepActiveInference2020,matsumoto2022goal}. In particular, several recent experiments posit active inference as a promising approach to optimal control and explainable and transparent artificial intelligence \cite{Friston2009,Friston2012,Sajid2021,mazzaglia2022free, Millidge2020a, albarracin2023designing}. In this article, we consider active inference as an approach to stochastic control, its current limitations, and how they can be overcome with dynamic programming and the adequate specification of a target distribution.

In the following three sections, we consider the active inference framework, discuss existing ideas accounting for perception,  planning and decision-making---and identify their limitations. Next, in Section \ref{dpefe_section}, we show how dynamic programming can address these limitations by enabling efficient planning and can scale up existing methods. We formalise these ideas in a practical algorithm for partially observed Markov decision processes (POMDP) in Section\ref{dpefe_algorithm}. Then we discuss the possibility of learning the agent's preferences by building upon Z-learning \citep{Todorov2006} in Section \ref{sec:learning_c}. We showcase these innovations with illustrative simulations in Section \ref{sec:simulations}.

\section{Active inference as biologically plausible optimal control}
\label{active_inference_intro}

The active inference framework is a formal way of modelling the behaviour of self-organising systems that interface with the external world and maintain a consistent form over time \cite{Friston2021,Kaplan2018,Kuchling2020}. The framework assumes that agents embody generative models of the environment they interact with, on which they base their (intelligent) behaviour \citep{Tschantz2020, Parr2018}. The framework, however, does not impose a particular structure on such models. Here, we focus on generative models in the form of partially observed Markov decision processes (POMDPs) for their simplicity and ubiquitous use in the optimal control literature \cite{Lovejoy1991,Shani2013,Kaelbling1998}. In the next section, we discuss the basic structure of POMDPs and how the active inference framework uses them. 

\subsection{Generative models using POMDPs}

Assuming agents have a discrete representation of their surrounding environment, we turn to the POMDP framework \citep{Kaelbling1998}. POMDPs offer a fairly expressive structure to model discrete state-space environments where parameters can be expressed as tractable categorical distributions. The POMDP-based generative model can be formally defined as a tuple of finite sets $(S, O, U, \mathbb{B},\mathbb{A})$:

\begin{itemize}
    \item [$\circ$] $s \in S:$ $S$ is a set of hidden states ($s$) causing observations $o$.
    \item [$\circ$]$o \in O:$ $O$ is a set of observations, where $o=s$, in the fully observable setting. In a partially observable setting,  $o=f(s)$.
    \item [$\circ$] $u \in U:$ $U$ is a set of actions ($u$) Eg: $U=\{\text{Left, Right, Up, Down} \}$. 
    \item [$\circ$]$\mathbb{B}:$ encodes the one-step transition dynamics, $P(s_{t} \vert s_{t-1}, u_{t-1})$ i.e., the probability that when action $u_{t-1}$ is taken while being in state $s_{t-1}$ (at time $t-1$) results in $s_{t}$ at time $t$.
    \item [$\circ$] $\mathbb{A}:$ encodes the likelihood mapping, $P(o_{\tau} \vert s_{\tau})$ for  the partially observable setting.
    \item [$\circ$] $\mathbb{D}:$ Encodes the prior of the agent about the hidden state factor $s$.
    \item [$\circ$] $\mathbb{E}:$ Encodes the prior of the agent about actions $u$.
    
\end{itemize}

In a POMDP, the hidden states ($s$) generate observations ($o$) through the likelihood mapping ($\mathbb{A}$) in the form of a categorical distribution, $P(o_{\tau} \vert s_{\tau}) = \mathrm{Cat}(\mathbb{A} \times s_\tau)$. 
$\mathbb{B}$ is a collection of square matrices $\mathbb{B}_{u}$, where $\mathbb{B}_{u}$ represents transition dynamics $P(s_{t} \vert s_{t-1}, u_{t-1} = u$): The transition matrix ($\mathbb{B}$) determines the dynamics of $s$  given the agent's action $u$ as $P(s_{t} \vert s_{t-1}, u_{t-1}) = \mathrm{Cat}(\mathbb{B}_{u_{t-1}} \times s_{t-1})$. In $\left[\mathbb{A} \times s_\tau \right]$ and $\left[ \mathbb{B}_{u_\tau} \times s_\tau \right]$, $s_\tau$ is represented as a one-hot vector that is multiplied through regular matrix multiplication \footnote{One-hot is a group of bits among which the legal combinations of values are only those with a single high (1) bit and all the others low (0). Here, the bit (1) is allocated to the state $s= s_\tau$ }. The \textit{Markovianity} of POMDPs means that state transitions are independent of history (i.e. state $s_{t}$ only depends upon the state-action pair $(s_{t-1}, u_{t-1})$ and not $s_{t-2}, ~u_{t-2}$ etc.).

In summary, the generative model can be summarised as follows,
\begin{equation}
P(o_{1:t},s_{1:t},u_{1:t}) = P(\mathbb{A}) P(\mathbb{B}) P(\mathbb{D}) P(\mathbb{E}) \prod_{\tau=1}^t P(o_{\tau} \vert s_{\tau}, \mathbb{A}) \prod_{\tau=2}^t P(s_{\tau} \vert s_{\tau-1}, u_{\tau-1},\mathbb{B}).
\end{equation}
So, from the agent's perspective, when encountering a stream of observations in time, such as $(o_{1}, o_{2}, o_{3}, ..., o_{t})$, as a consequence of performing a stream of actions $(u_{1}, u_{2}, u_{3}, ..., u_{t-1})$, the generative model quantitatively couples and quantifies the causal relationship from action to observation through some assumed hidden states of the environment. These are called `hidden' states because, in POMDPs, the agent cannot observe them directly. Based on this representation, an agent can now attempt to optimise its actions to keep receiving preferred observations. 
Currently, the generative model has no concept of `preference' and `goal' \citep{bruineberg2018free}. Rather than attempting to maximise cumulative reward from the environment, active inference agents minimise the `surprise' of encountered observations \citep{Sajid2021,sajid2021exploration}. We look at this idea closely in the next section.

\subsection{Surprise and free energy}

The surprise of a given observation in active inference \citep{Friston2019, Sajid2021} is defined through the relation
\begin{equation}\label{eq:surprise}
    S(o) = -\text{log}(P(o)).
\end{equation}
Please note that the agent does not have access to the true probability of an observation: $P_{\text{true}}(o)$. However, the internal generative model expects an observation with a certain probability $P(o)$, which quantifies surprise in Eq.\ref{eq:surprise}. Minimising surprise directly requires the marginalisation of the generative model, i.e., $P(o) = \sum_{s} P(o,s)$, which is often computationally intractable due to the large size of the state-space \citep{bleiVariationalInferenceReview2017,sajid2022bayesian}. Since $f(x) = \text{log}(x)$ is a convex function, we can solve this problem by defining an upper-bound to surprise using Jensen's inequality \footnote{Jensen's inequality: If $X$ is a random variable and $\psi$ is a convex function, $\psi \left( \mathbb{E}[X]\right) \leq \mathbb{E}\left[ \psi(X) \right]$.}:

\begin{equation}\label{eq:definingF}
    S(o)  = -\text{log}\sum_{s} P(o,s) \leq -\sum_{s} Q(s)\text{log}\frac{P(o,s)}{Q(s)} = F[Q].
\end{equation}

The newly introduced term $Q(s)$ is often interpreted as an (approximate posterior) belief about the (hidden) state: $s$. This upper bound (F) is called the variational free energy (VFE) (it is also commonly known as evidence lower bound -- ELBO \citep{bleiVariationalInferenceReview2017} \footnote{The connection between the two lies in the fact that they are essentially equivalent up to a constant (the log evidence), but with opposite signs. In other words, minimizing VFE is equivalent to maximizing the ELBO. Formally this is: 
$VFE = - ELBO + constant$}). So, by optimising the belief $Q(s)$ to minimise the variational free energy (F), an agent is capable of minimising the surprise $S(o)=-\text{log}(P(o))$ or at least maintain it bounded at low values.
  
\paragraph{How is this formulation useful for stochastic control?} Imagine the agent embodies a biased generative model with `goal-directed' expectations for observations. The goal then becomes to minimise $F$, which can be done through the conjunction of perception, i.e., optimising the belief $Q(s)$, or action, i.e., controlling the environment to sample observations that lead to a lower $F$ \citep{Tschantz2020}. So, instead of passively inferring what caused observations, the agent starts to `actively' infer, exerting control over the environment using available actions in $U$. The central advantage of this formalism is that there is now only one single cost function ($F$) to optimise all aspects of behaviour, such as perception, learning, planning, and decision-making (or action selection). There are related works in the reinforcement literature noting the use of similar information-theoretic metrics for control \cite{rhinehart2021information, berseth2019smirl}. The following section discusses this feature in detail and further develops the active inference framework.

\section{Perception and learning}
\label{perception_and_learning}

\subsection{Perception}

From the agent's perspective, perception means (Bayes optimally) maintaining a belief about hidden states $s$ causing the observations $o$. In active inference, agents optimise the beliefs $Q(s)$ to minimise $F$. The VFE may be rewritten (from Eq.\ref{eq:definingF}), using the identity $P(o,s) = P(s)P(o \vert s)$, as:

\begin{equation}\label{eq1_perception}
    F = \sum_{s} Q(s)[~\text{log} Q(s)-\text{log} P(o\vert s) - \text{log}P(s)~].
\end{equation}

Differentiating $F$ w.r.t $Q(s)$ and setting the derivative to zero, we get (see Supplementary \ref{appendix:perception_derivation}),
\begin{equation}\label{eq2_perception}
    \frac{\delta F}{\delta Q(s)} = \sum_{s} 1 + \text{log}Q(s) - \text{log}P(o\vert s) - \text{log}P(s) = 0.
\end{equation}

Using the above equation, we can evaluate the optimal $Q(s)$ that minimises\footnote{The second derivate of Eq.\ref{eq2_perception} w.r.t to $Q(s)$ is greater than zero which corresponds to local minima of $F$ w.r.t to $Q(s)$.} $F$ using,
\begin{equation}\label{eq3_perception}
    \text{log}Q^*(s) = \text{log}P(s) +\text{log}P(o\vert s).
\end{equation}

This equation provides the (Bayesian) belief propagation scheme, given by
\begin{equation}\label{beliefspomdp}
\underbrace{Q(s_{t+1})}_\text{Posterior} = \sigma \left(\underbrace{\text{log}P(s_{t+1})}_\text{Prior} + \underbrace{\text{log}(o_{t+1} \cdot \mathbb{A} s_{t+1})}_\text{Likelihood} \right).
\end{equation}

Here, $\sigma$ is the softmax function; that is, the exponential of the input that is then normalised so that the output is a probability distribution. Given a real-valued vector $V$ in $\mathbb{R}^K$, the i-th element of $\sigma(V)$ reads:
\begin{equation}
    \sigma(V)^i = \frac{\exp{V^{i}}}{\sum_{j=1}^K \exp{V^{j}}},
\end{equation}

where $V^{i}$ corresponds to the i-th element of $V$.
We estimate the first term of Eq.\ref{beliefspomdp}, i.e. the prior using belief $Q(s_{t})$ at time $t$, and the action $u_t$ taken at time $t$. Using the transition dynamics operator $\mathbb{B}_{u_t}$, we write:

\begin{equation}\label{eq:prior}
    P(s_{t+1})=\mathbb{B}_{u_t} \cdot Q(s_{t}).
\end{equation}

At the first time step, i.e. $t=0$, we use a known prior about the hidden state $\mathbb{D}$ to substitute for the term $P(s_{t+1})$.
Similarly, the second term in Eq.\ref{beliefspomdp}, i.e., the estimate of the hidden state from the observation we gathered from the environment at time $t+1$ can be evaluated as the dot product between the likelihood function $\mathbb{A}$ and the observation gathered at time $t+1$. The belief propagation scheme here is shown in the literature to have a degree of biological plausibility in the sense that it can be implemented by a local neuronal message-passing scheme \citep{Vries2017}. The following section discusses the learning of the model parameters.

\subsection{Learning}

The parameter learning rules of our generative model are defined in terms of the optimised belief about states $Q(s)$. 
In our architecture, the agent uses belief propagation \footnote{We stick to the belief propagation scheme for perception in this paper. However, general schemes like variational message passing may be used to estimate $Q(s)$.} to best estimate $Q(s)$, the belief about (hidden) states in the environment. Given these beliefs, observations sampled, and actions undertaken by the agent, the agent hopes to learn the underlying contingencies of the environment. The learning rules of active inference consist of inferring parameters of $\mathbb{A}$, $\mathbb{B}$, and $\mathbb{D}$ at a slower time scale. We discuss such learning rules in detail in the following.

\subsubsection{Transition dynamics} \label{learningmdp}

Agents learn the transition dynamics, $\mathbb{B}$, across time by maintaining a concentration parameter $b_{u}$, using conjugate update rules well documented in the active inference literature\citep{Friston2017,DaCosta2020,Sajid2021} such as:

\begin{equation}
    b_u \gets b_u + Q(u_{t-1}) \cdot \left(Q(s_{t}) \otimes Q(s_{t - 1}) \right),
    \label{eqn:learning}
\end{equation}

where $Q(u)$ is the probability of taking action $u$, $Q(s_{t})$ is belief the state at time $t$ as a consequence of action $u$ at $t-1$, and $Q(s_{t}) \otimes Q(s_{t - 1})$ represents a square matrix of Kronecker product between two vectors $Q(s_{t})$ and $Q(s_{t-1})$.

Every column of the transition dynamics $\mathbb{B}_u$, can be estimated from $b_u$ column-wise as,

\begin{equation}
    \text{col}(\mathbb{B}_{u})_{i} = \overline{Dir} \left[ \text{col}(b_{u})_{i} \right].
\end{equation} 

Here, $\text{col}(X)_{i}$ is the i-th column of $X$. $\overline{Dir}(b_{u})$ represents the mean of the Dirichlet distribution \footnote{Dirichlet distributions are commonly used as prior distributions in Bayesian statistics given that the Dirichlet distribution is the conjugate prior of the categorical distribution and multinomial distribution. We use the mean of the Dirichlet distribution here because we are performing a Bayesian update. Briefly, the mean of a Dirichlet distribution with parameters $\boldsymbol{b} = (b_1, ..., b_K)$ is given by $\boldsymbol{\mu} = (\mu_1, ..., \mu_K)$ where $\mu_k = \frac{b_k}{\sum_{j=1}^{K} b_j}$. So, in this case, each entry in the estimated transition probabilities is the corresponding entry in $b_u$, divided by the sum of all entries in the corresponding column of $b_u$. This normalization ensures that the columns of $\mathbb{B}_u$ sum to 1.} with parameter $b_{u}$.

\subsubsection{Likelihood}

Similar to the conjugacy update in Eq.\ref{eqn:learning}, the Dirichlet parameter ($a$) for the likelihood dynamics ($\mathbb{A}$) is learned over time within trials using the update rule,
\begin{equation}\label{eqn:liklihood}
    a \gets a + o_{t} \otimes Q(s_{t}).
\end{equation}

Here, $o_{t}$ is the observation gathered from environment at time $t$, and $Q(s_{t})\approx P(s_{t} \vert o_{1:t})$ is the approximate posterior belief about the hidden-state ($s$) \citep{Friston2017,DaCosta2020}. 

Like perception and learning, decision-making and planning can also be formulated around the cost function $F$ and belief $Q$. In the next section, we review in detail existing ideas \citep{Friston2021,sajid2022active} for planning and decision-making. We then identify their limitations and, next, propose an improved architecture. 

\section{Planning and decision making}
\label{planning_decision_making}
\subsection{Classical formulation of active inference}

Traditionally, planning and decision-making by active inference agents revolve around the goal of minimising the variational free energy of the observations one expects in the future. To implement this, we define a policy space comprising sequences of actions in time. The policy space in classical active inference \citep{Sajid2021} is defined as a collection of policies $\pi_n$

\begin{equation}
    \Pi = \{\pi_{1}, \pi_{2}, ..., \pi_{N}\},
\end{equation}
which are themselves sequences of actions indexed in time; that is, $\pi = (u_{1}, u_{2}, ..., u_{T})$, where $u_t$ is one of the available action in $U$, and $T$ is the agent's planning horizon. $N$ is the total number of unique policies defined by permutations of available actions $u$ over a time horizon of planning $T$.

To enable goal-directed behaviour, we need a way to quantify the agent's preference for sample observations $o$. The prior preference for observations is usually defined as a categorical distribution over observations,
\begin{equation}
    \mathbb{C} = \mathrm{Cat}(o).
\end{equation}

So, if the value corresponding to an observation in $\mathbb{C}$ is the highest, it is the most preferred observation for the agent. 
Given these two additional parameters ($\Pi$ and $\mathbb{C}$), we can define a new quantity called the expected free energy (EFE) of a policy $\pi$ similar to the definition in \citep{Sajid2021,schwartenbeck2019computational,Parr2019a} as,

\begin{equation}\label{EFE_CAI_eqn}
    G(\pi) = \sum_{t=1}^T \underbrace{D_{KL}\left[ Q(o_{t} \vert \pi^t)  ~\vert \vert~ \mathbb{C} \right]}_\text{Risk} + \underbrace{\mathbb{E}_{Q(s_{t} \vert s_{t-1}, \pi^{t-1})} \left[ \mathbb{H}\left[ P(o_{t} \vert s_{t})\right] \right]}_\text{Expected ambiguity}.
\end{equation}

In Eq.\eqref{EFE_CAI_eqn} above, $\pi^t$ is the t-th element in $\pi$, i.e. the action corresponding to time $t$ for policy $\pi$. The term, $Q(o_{t} \vert \pi^t)$ represents the most likely observation caused by the policy $\pi$ at time $t$. $D_{KL}$ stands for the KL-divergence, which, when minimised, forces the distribution $Q(o_{t} \vert \pi^t)$ closer towards $\mathbb{C}$. This term is also called the "Risk" term, representing the goal-directed behaviour of the agent. The KL-divergence between two distributions, $P$ and $Q$, is defined as:
\begin{equation}
    D_{KL}(P \vert \vert Q) = \sum_{i} P(i) \text{log} \frac{P(i)}{Q(i)},
\end{equation}
and $P=Q$ if and only if $D_{KL}(P \vert \vert Q) = 0$.

In the second term of Eq.\ref{EFE_CAI_eqn}, $\mathbb{H}\left[ P(o_{t} \vert s_{t}) \right]$ stands for the (Shannon) entropy of $P(o_{t} \vert s_{t})$ defined as, 
\begin{equation}
    \mathbb{H}(P(o)) = -\sum_{o \epsilon O} P(o)\text{log}P(o).
\end{equation}

The second term is also called the `Expected ambiguity' term. When the expected entropy of $P(o_{t} \vert s_{t})$ w.r.t the belief $Q(s_{t} \vert s_{t-1}, \pi^{t-1})$ is less, the agent is more confident of the state-observation mapping (i.e., $\mathbb{A}$) in its generative model. 

Hence, by choosing the policy $\pi$ to make decisions that minimise $G$, the agent minimises `Risk' at the same time and also its `Ambiguity' about the state-observation mapping. Hence, in active inference, decision-making naturally balances the exploration-exploitation dilemma \citep{TRICHE202216}. We also note that the agent is not optimising $G$ but only evaluating and comparing various $G$ over different policies $\pi$ in the policy space $\Pi$. Once the best policy $\pi$ is identified, the most simple decision-making rule follows by choosing actions $u_{t} = \pi^t$ at time $t$, where $\pi^t$ is the $t$-th element of $\pi$.

It may already be evident that the above formulation has one fundamental problem: in the stochastic control problems that are commonly encountered in practice, the size of possible action spaces $U$ and the time horizons of planning $T$ make the policy space too large to be computationally tractable. For example, with eight available actions in $U$ and a time horizon of planning $T=15$, the total number of (definable) policies that need to be considered are ($3.5*10^{13}$) $35$ trillion. Even for this relatively small-scale example, this policy space is not computationally tractable to simulate agent behaviour (unless additional decision tree search methods are considered \citep{fountasDeepActiveInference2020,championBranchingTimeActive2021a,championBranchingTimeActive2021} or policy amortisation \cite{fountasDeepActiveInference2020,Catal2020}) or by eliminating implausible policy trajectories using Occam's principle. We now turn to an improved scheme that redefines policy space and planning all together.

\subsection{Sophisticated inference}

Graduating from the classical definition of policy as a sequence of actions in time, sophisticated inference \citep{Friston2021} attempts to evaluate the EFE of observation-action pairs at a given time $t$, $G(o_{t}, u_{t})$. Given this joint distribution, an agent can sample actions using the conditional distribution $Q(u_t \vert o_{t})$ when observing $o_{t}$ at time $t$,

\begin{equation}
    u_{t} \sim Q(u_t \vert o_{t}) = \sigma \left[ -G(o_{t}, u_{t})\right].
\end{equation}

Given the prior-preference distribution of an agent $P(s)$, in terms of hidden states $s$, the expected free energy of an observation-action pair is defined as \citep{Friston2021},

\begin{align}\label{SIAF_equation}
    G(o_{t}, u_{t}) = \underbrace{\mathbb{E}_{P(o_{t+1} \vert s_{t+1}) Q(s_{t+1} \vert u_{<t+1})} \left[ \underbrace{\text{log} Q(s_{t+1} \vert u_{<t+1}) - \text{log}P(s_{t+1})}_\text{Risk} \underbrace{ - \text{log} P(o_{t+1} \vert s_{t+1})}_\text{Ambiguity} \right]}_\text{EFE of action at time $t$} + \\ \underbrace{\mathbb{E}_{Q(u_{t+1} \vert o_{t+1})Q(o_{t+1} \vert u_{\leq t})}\left[ G(o_{t+1}, u_{t+1}) \right]}_\text{EFE of future actions}.
\end{align}

We rewrite this equation in a familiar fashion to Eq.\ref{EFE_CAI_eqn}. In the above equation, the agent holds evaluated beliefs about future hidden states given all past actions in the term $Q(s_{t+1} \vert u_{<t+1})$. Beliefs about hidden states can be extrapolated to observations using the likelihood mapping ($\mathbb{A}$) as

\begin{equation}\label{so_mappingeqn}
    P(o_{t+1} \vert s_{t+1}) Q(s_{t+1} \vert u_{<t+1}) = Q(o_{t+1} \vert u_{<t+1}).
\end{equation}

Also, the prior preference of the agent is defined in terms of hidden states $s$ in Eq.\ref{SIAF_equation}. Now the Eq.\ref{SIAF_equation} can be rewritten using mappings like Eq.\ref{so_mappingeqn} as,

\begin{align}\label{SIAF_equation_2}
    G(o_{t}, u_{t}) = \underbrace{\mathbb{E}_{Q(o_{t+1} \vert u_{<t+1})} \left[ \underbrace{\log Q(o_{t+1} \vert u_{<t+1}) - \log\mathbb{C}}_\text{Risk} \underbrace{ - \log P(o_{t+1} \vert s_{t+1})}_\text{Ambiguity} \right]}_\text{EFE of action at time $t$} + \\ \underbrace{\mathbb{E}_{Q(u_{t+1} \vert o_{t+1})Q(o_{t+1} \vert u_{\leq t})}\left[ G(o_{t+1}, u_{t+1}) \right]}_\text{EFE of future actions}.
\end{align}

Note that the prior preference distribution in the equation above is over observations $o$, $\mathbb{C}=P(o)$.
Rewriting Eq.\ref{SIAF_equation_2} in a similar fashion to the previously discussed classical active inference we obtain

\begin{align}\label{SIAF_equation_3}
    G(o_{t}, u_{t}) = \underbrace{\underbrace{ D_{KL} \left[ Q(o_{t+1} \vert u_{<t+1}) ~\vert \vert~ \mathbb{C} \right] }_\text{Risk} + \underbrace{\mathbb{E}_{Q(s_{t+1} \vert u_{<t+1})} \mathbb{H} \left[ P(o_{t+1} \vert s_{t+1}) \right]}_\text{Expected ambiguity}}_\text{EFE of action at time t} + \\ \underbrace{\mathbb{E}_{Q(u_{t+1} \vert o_{t+1})Q(o_{t+1} \vert u \leq t)}\left[ G(o_{t+1}, u_{t+1}) \right]}_\text{EFE of future actions}.
\end{align}

\begin{figure}[t!]
    \centering
    \includegraphics[width=\textwidth]{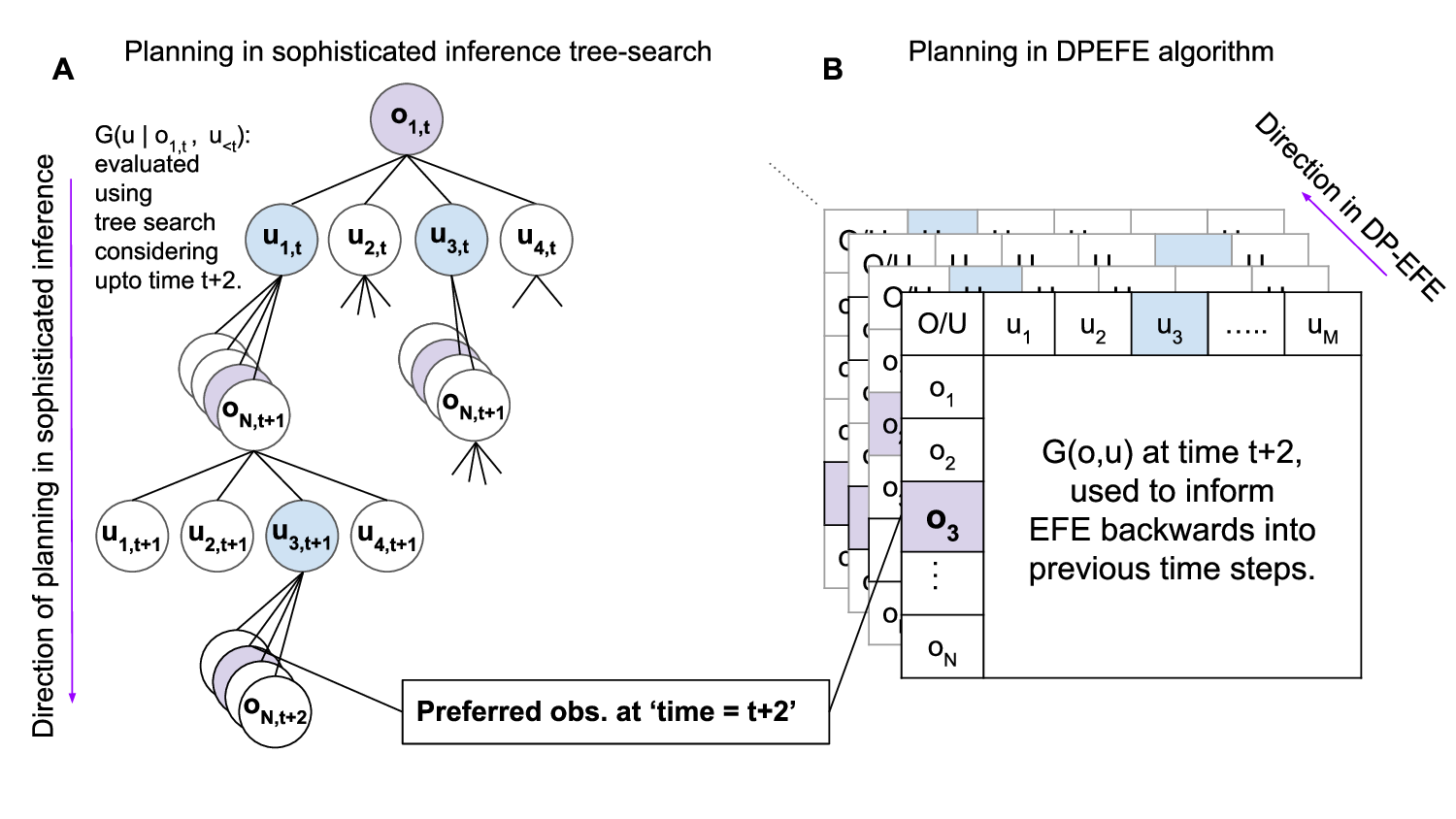}
    \caption{Graphics to compare and contrast the differences between the sophisticated inference and DPEFE (Dynamic programming in expected free energy) algorithm planning schemes. A: Sophisticated inference algorithm uses an extensive tree search, going forward in time, to accumulate free energy of the future paths. So, an agent's preference for observations, when matched with future predictions, will inform an optimal state-action trajectory, as shown in the tree search. Light-purple states represent the preferred observations at that given time step, and light-blue actions are the optimal actions inferred through the tree search. As noted in \cite{Friston2021}, an agent can significantly reduce the tree search complexity by terminating the search when the action probability falls below a certain threshold. However, this approximation does not guarantee optimal policy as the agent might miss preferred observations deeper in the tree search. B: In the DPEFE algorithm, an agent starts planning backwards from a fixed planning horizon. Here, the EFE of future states informs EFE of state-action pairs one step backward in time. Hence, the planning complexity of tree search is avoided, but the preference for future states propagates to influence decisions at previous time steps. Since the agent needs to evaluate only a table (of EFE) at every planning step, this planning algorithm is linear in time, number of states, and number of actions.}
    \label{fig:dpefevssi}
\end{figure}

The first two terms can be interpreted the same way as we did for Eq.\ref{SIAF_equation_2} in the previous section. However, the third term in Eq.\ref{SIAF_equation_3} gives rise to a recursive tree-search algorithm, accumulating free energies of the future (as deep as we evaluate forward in time). Such an evaluation is pictorially represented in Fig.\ref{fig:dpefevssi} (A).

While Bellman optimal \cite{DaCosta2021}, one unavoidable limitation of the sophisticated inference planning algorithm is that it faces a worse curse of dimensionality for even relatively small planning horizons. For example, to evaluate the goodness of an action within a period of fifteen time-steps into the future, and with eight available actions and a hundred hidden states, requires an exorbitant $(100*8)^{15}$ ($\approx 3.5*10^{43}$) calculations, in comparison to $100*(8)^{15}$($\approx 3.5*10^{15}$) for classical active inference. A simple solution proposed in \citep{Friston2021} is to eliminate tree search branches by setting a threshold value to predictive probabilities such as $Q(u_{t+1} \vert o_{t+1})$ in Eq.\ref{SIAF_equation_3}.
So, for example, when $Q(u_{t+1} \vert o_{t+1}) < 1/16$ during planning, the algorithm terminates the search over future branches. This restriction significantly reduces the computational time, and a set of ensuing (meaningful) simulations was presented in \citep{Friston2021}.

\begin{figure}
    \centering
    \includegraphics[width=\textwidth]{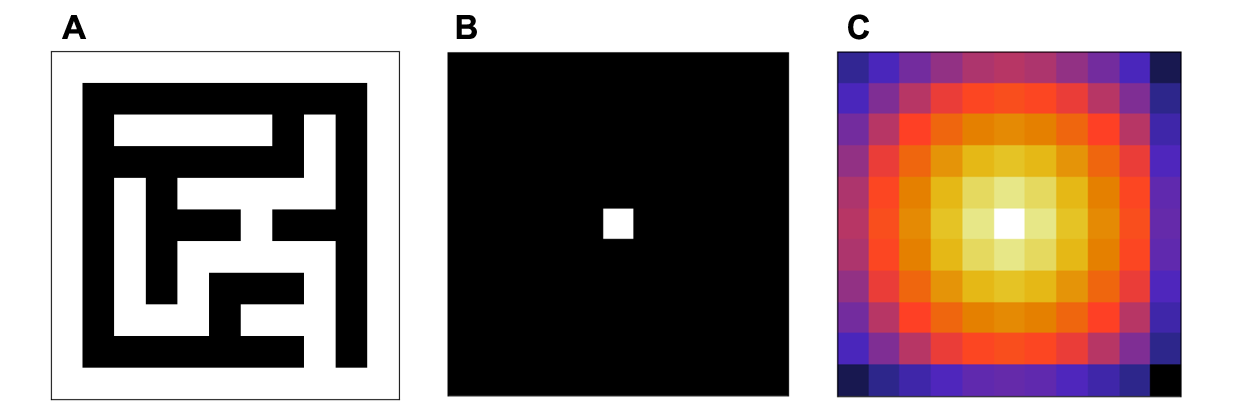}
    \caption{Informed and uninformed prior preferences: A: A navigation problem, B: A strictly defined, sparse prior preference which has information only about the final goal state, C: Informed prior preference necessary for `pruning of tree search' in sophisticated inference (light colour states are more preferred)}\label{fig:prior_saif}
\end{figure}

Another limitation is that in all active or sophisticated inference agents to facilitate desirable behaviour, a prior preference needs to be defined by the modeller or learned by the agents \cite{sajid2021exploration,sajid2022modelling} informing the agent that some states are preferable to others, as demonstrated in Fig.\ref{fig:prior_saif} (B) for a grid problem given in Fig.\ref{fig:prior_saif} (A).  An informed prior preference enables the agent to solve this navigation task by only planning four or more time steps ahead. It can take action and move towards a `more preferred state' if not the final goal state. However, without such information, the agent is `blind' (cf. Fig.\ref{fig:prior_saif} (C)) and can only find the optimal move when planning the whole eight-step trajectory for the given grid.

We first noticed this limitation when comparing different active inference schemes to various well-known reinforcement learning algorithms in \citep{Paul2021} in a fully observable setting (i.e., MDPs). In the next section, we demonstrate how to scale the sophisticated inference scheme using dynamic programming for the general case of POMDP-based generative models.

\section{Dynamic programming for evaluating expected free energy}
\label{dpefe_section}

The Bellman optimality principle states that the sub-policy of an optimal policy (for a given problem) must itself will be an optimal policy for the corresponding sub-problem \citep{Sutton2018}. Dynamic programming is an optimisation technique that naturally follows the Bellman optimality principle; rather than attempting to solve a problem as a whole, dynamic programming attempts to solve sub-parts of the problem and integrate the sub-solutions into a solution to the original problem. This approach makes dynamic programming scale favourably, as we solve one sub-problem at a time to integrate later. The more we break down the large problem into corresponding sub-problems, the more computationally tractable is the solution.

Inspired by this principle, let us consider a spatial navigation problem that the agent needs to solve in our setting. The optimal solution to this navigation problem is a sequence of individual steps. Our prior preference for the `goal state' is for the end of our time horizon of planning. So, the agent may start planning from the last time step (a step sub-problem) and go backwards to solve the problem. This approach is also called planning by backward induction \cite{zhou2019collaborative}.

So, for a planning horizon of $T$ (i.e., the agent aims to reach goal state at time $T$), the EFE of the (last) action for the $T-1$th time step in a POMDP setting is written as:

\begin{equation} \label{efeeqnpomdp}
    G(u_{T-1}, o_{T-1}) = D_{KL} [Q(o_{T} ~\vert~ u_{T-1},s_{T-1}) ~\vert \vert~ \mathbb{C}].
\end{equation}

The term, $G(u_{T-1} \vert o_{T-1})$ is the expected free energy associated with any action $u_{T-1}$, given we are in (hidden) state  $s_{T-1}$. This estimate measures how much we believe observations at time $T$ will align with our prior preference $\mathbb{C}$.
Note that, for simplicity, we ignored the `expected ambiguity' term in the equation above, i.e. the uncertainty of state-observation mapping (or likelihood), cf. Eq.\ref{SIAF_equation_3}. This does not affect our subsequent derivations; we can always add it as an additional term. The following derivation provided technical details of dynamic programming while focusing only on the `risk' term in $G$.

To estimate $Q(o_{T} \vert u_{T-1},s_{T-1})$, we make use of the prediction about states $Q(s_{T})$ that can occur at time $T$:

\begin{equation}\label{eq_g}
    Q(s_{T}~ \vert~ u_{T-1} ,s_{T-1})=\mathbb{B}_{u_{T-1}} \cdot Q(s_{T-1}),
\end{equation}

and given the prediction $Q(s_{T})$, we write 
\begin{align}\label{Eq.obs}
    Q(o_{T} ~\vert~ u_{T-1},s_{T-1})=\mathbb{A} \cdot Q(s_{T}~ \vert~ u_{T-1} ,s_{T-1}) \\
    = \mathbb{A} \cdot \left( \mathbb{B}_{u}^{T-1} \cdot Q(s_{T-1}) \right).
\end{align}

Next, using Eq.\ref{eq_g}, the corresponding action distribution (for action selection) is calculated at time $T$,
\begin{equation}\label{actdist}
    Q(u_{T-1} \vert o_{T-1})=\sigma \left( - G(u_{T-1} \vert o_{T-1}) \right),
\end{equation}

where we recursively calculate the expected free energy for actions and the corresponding action-distributions for time steps $T-2, T-3, ..., t=1$ backwards in time,\footnote{For times other than $T-1$, the first term in Eq.\ref{eq.rec.pl} does not contribute to solving the particular instance if $\mathbb{C}$ only accommodates preference to a (time-independent) goal-state. However, for a temporally informed $\mathbb{C}$, i.e. with a separate preference for reward maximisation at each time step, this term will meaningfully influence action selection.}

\begin{equation}\label{eq.rec.pl}
    G(u_{t} \vert o_{t})= \underbrace{D_{KL} [Q(o_{t+1} ~\vert~ u_{t},s_{t}) ~\vert \vert~ \mathbb{C}]}_\text{EFE of action at time $t$} + \underbrace{\mathbb{E}_{Q(o_{t+1},u_{t+1} \vert o_{t},u_{t})}[ G(u_{t+1} \vert o_{t+1})]}_\text{EFE of next action at $t+1$}.
\end{equation}

In the equation above, the second term condenses information about all future observations rather than doing a forward tree search in time. 
To inform $G(u_{t} \vert o_{t})$, we consider all possible observation-action pairs that can occur in time $t+1$ and use the previously evaluated $G(u_{t+1} \vert o_{t+1})$.
In Eq.\ref{eq.rec.pl}, we evaluate $Q(o_{t+1},u_{t+1} \vert o_{t},u_{t})$ using,
\begin{equation}\label{eq_q-split}
    Q(s_{t+1},u_{t+1} \vert s_{t},u_{t})= \underbrace{Q(s_{t+1} \vert s_{t},u_{t})}_\text{$\mathbb{B}$} \cdot \underbrace{Q(u_{t+1} \vert s_{t+1})}_\text{Action distribution}.
\end{equation}
We then map the distribution $Q(s_{t+1},u_{t+1} \vert s_{t},u_{t})$ to the observation space and evaluate $Q(o_{t+1},u_{t+1} \vert o_{t},u_{t})$ using the likelihood mapping $\mathbb{A}$.
In Eq.\ref{eq_q-split}, we assume that actions in time are independent of each other, i.e. $u_{t}$ is independent of $u_{t+1}$. Even though actions are assumed to be explicitly independent in time, the information (and hence desirability) about actions are also informed backwards in time from the recursive evaluation of expected free energy. 

While evaluating the EFE, $G$, backwards in time, we used the action distribution in Eq.\ref{actdist}. This action distribution can be directly used for action selection.
Given an observation $o$ at time $t$, $u_{t}$ may be sampled \footnote{Precision of action-section may be controlled by introducing a positive constant inside the softmax function $\sigma(.)$ in Eq.\ref{actdist}. The higher the constant, the higher the chance of selecting the action with less EFE.} from, 
\begin{equation}\label{eqn:action_sampling}
    u_{t} \sim Q(u_{t} \vert o_{t} = o).
\end{equation}

In the next section, we summarise the above formulation as a novel active inference algorithm useful for modelling intelligent behaviour in sequential POMDP settings.

\subsection{Algorithmic formulation of DPEFE}
\label{dpefe_algorithm}

Here, we formalise a generic algorithm that can be employed for a sequential POMDP problem. The main algorithm (see Alg.\ref{alg:main}) works sequentially in time and brings together three different aspects of the agent's behaviour, namely, perception (inference), planning, and learning. 

For planning, that is, to evaluate the expected free energy ($G$) for actions (given states) in time, we employ the planning algorithm (See. Alg.\ref{alg.planning}) as a subroutine to Alg.\ref{alg:main}. In the most general case, the algorithm is initialised with 'flat' priors for the likelihood function ($\mathbb{A}$) and transition dynamics ($\mathbb{B}$). The algorithm also allows us to equip the agent with a more informed prior about $\mathbb{A}$ and $\mathbb{B}$. Learning $\mathbb{C}$ in the DPEFE algorithm is setting $\mathbb{C}$ as a one-hot vector with the encountered goal state. This technique accelerates the parameters' learning process during trials and improves agent performance. We can also make available the `true' dynamics of the environment to the agent whenever present. With `true' dynamics available at the agent's disposal, the agent can infer hidden states and plan accurately.
The next section discusses a different approach to ameliorating the curse of dimensionality in sophisticated inference. Later, we discuss a potential learning rule for the prior preference distribution $\mathbb{C}$ inspired by a seminar work in the control theoretic literature.

\begin{algorithm}[!th]
\caption{Active inference in a sequential POMDP}\label{alg:main}
\begin{algorithmic}

\State $\mathbb{C} \gets o_{goal}$ \Comment{prior preference (Known / Learned as one-hot vector (sparse) when encountering of goal-state}
\State $\mathbb{D} \gets s_{start}$ \Comment{Known/Learned at time $t=1$}
\State $T \gets$ Planning horizon of the DPEFE agent
\State $a \gets a_{prior}+\epsilon$ \Comment{The prior is usually an uninformed 'flat' distribution.}
\State $\mathbb{A} \gets \overline{Dir}(a)$
\State $b_u \gets b_{prior}+\epsilon ~\forall~ u$ \Comment{$\epsilon$ is a negligible positive value to ensure numerical stability}
\State $\mathbb{B}_{u} \gets \overline{Dir}(b_{u}) ~\forall~ u$
\State $T_{max}$ \Comment{Threshold for episode length set in Environment}
\State
\While {True} \Comment{Loop forever}
\For{$t$ from 1 to $T_{max}$}
\State
\State \textbf{Inference:}
\If{$t=1$}
    \State {$P(s_{1}) \gets \mathbb{D}$ (known)} \Comment{Prior at $t=1$}
    \State Observe $o_{1}=o_{start}$ from Environment
    \State Evaluate, $Q(s_{t=1})$ \Comment{Inference, Ref Eq.\ref{beliefspomdp}}
\Else
    \State Evaluate $P(s_{t})$ \Comment{Ref. Eq.\ref{eq:prior}}
    \State Evaluate $Q(s_{t})$ \Comment{Inference, Ref Eq.\ref{beliefspomdp}}
\EndIf
\State
\State \textbf{Planning and action selection}
\State Evaluate $G(u_{t} \vert o_{t})$ ~ $\forall ~ t ~\epsilon~ 1,..,T-1,~ o ~\epsilon~ O$ \Comment{Planning, Ref. Algorithm.\ref{alg.planning}}
\State Evaluate $Q(u_{t})$ ~ $\forall ~ t, o$ \Comment{Action distribution, Ref. Eq.\ref{actdist}}
\State Sample $u_{t} \sim Q(u_{t})$. \Comment{Sample action, Ref Eq.\ref{eqn:action_sampling}}
\State Observe $o_{t+1} \gets$ From-Environment by taking action $u_{t}$
\State
\State \textbf{Learning}
\If{$t>1$}
    \State $ b_u \gets b_u + Q(u_{t-1}) \cdot \left(Q(s_{t}) \otimes Q(s_{t - 1}) \right) $ \Comment{Ref Eqn.\ref{eqn:learning}}
\EndIf
\State $a \gets a +  o_{t} \otimes Q(s_{t})$ \Comment{Ref Eqn.\ref{eqn:liklihood}}
\State
\State \textbf{End of trial}
\If{$t=T_{max}$}
\State Environment is reset i.e $s_{True} \gets s_{start}$
\State $t \gets$ 1
\EndIf
\If{\textit{Goal-achieved}}
\State $\mathbb{C} \gets \textit{One-hot}(o_{goal}$)
\State Environment is reset i.e $s_{True} \gets s_{start}$
\State $t \gets$ 1
\EndIf
\EndFor
\State
\State $\mathbb{A} = \overline{Dir}(a)$ \Comment{Updating likelihood}
\State $\mathbb{B}_{u} = \overline{Dir}(b_{u})$ \Comment{Update beliefs about transition dynamics}
\EndWhile \Comment{End of experiment}
\end{algorithmic}
\end{algorithm}

\begin{algorithm}
\caption{Planning backwards in time}\label{alg.planning}
\begin{algorithmic}
\State $\mathbb{A} \gets$ Passed from Alg.\ref{alg:main}
\State $\mathbb{B} \gets$ Passed from Algorithm.\ref{alg:main}
\State $\mathbb{C} \gets$ Passed from Algorithm.\ref{alg:main}
\State $T \gets$ Passed from Algorithm.\ref{alg:main}
\State
\State \textbf{Planning}
\For{$t=T-1$ to $t=1$}
    \If{$t=T-1$}
    \State Evaluate $Q(o_{T} ~\vert~ u_{T-1},s_{T-1})$ \Comment{Ref. Eq.\ref{Eq.obs}}
    \State Evaluate $G(u_{T-1} \vert o_{T-1})$ \Comment{Ref. Eq.\ref{efeeqnpomdp}}
    \Else
    \State Evaluate $Q(o_{t} ~\vert~ u_{t-1},o_{t-1})$ \Comment{Ref. Eqn.\ref{Eq.obs}}
    \State Evaluate $G(u_{t} \vert o_{t})$ \Comment{Ref. Eq.\ref{eq.rec.pl}}
    \EndIf
\EndFor
\end{algorithmic}
\end{algorithm}

\section{Learning prior preferences}
\label{sec:learning_c}

In the previous section, we introduced a practical algorithm solution that speeds up planning in sophisticated inference. The second innovation on offer is to enable learning of preferences $\mathbb{C}$ such that smaller planning horizons become sufficient for our agent to take optimal actions, as discussed in Fig.\ref{fig:prior_saif}. 
A seminal work from the literature on control theory proposes using a `desirability' function, scoring how desirable each state is, to compute optimal actions for a particular class of MDPs and, importantly, showing that the planning complexity of computing those actions is linear in time \citep{Todorov2006}. When the underlying MDP model of the environment is unavailable and the agent needs to take actions based solely on a stream of samples of states and rewards (i.e., $s_{t}, r_{t}, s_{t+1}$), an online algorithm called Z-learning, inspired from the theoretical developments in \citep{Todorov2006}, was proposed to solve this problem.
Given an optimal desirability function $z(s)$, the optimal control, or policy, is analytically computable. The calculation of $z(s)$ does not rely on knowledge of the underlying MDP but instead, on the following online learning rule:
\begin{equation}\label{eq:z_learning_rule}
    \hat{z}(s_{t}) \leftarrow (1 - \eta_{t})\hat{z}(s_{t}) + \eta_{t}~exp(r_{t})~\hat{z}(s_{t+1}),
\end{equation}
where, $\eta$ is a learning rate that is continuously optimised---see below. These two terms form a weighted average that updates the estimate of $\hat{z}(s_t)$, with $\eta_t$ controlling the balance between the old estimate and the new information.

Inspired by these developments, we write a learning rule for updating $\mathbb{C}$ which can be useful for the sophisticated inference agent. Given the samples $(o_{t}, r_{t}, o_{t+1})$, an agent may learn the parameter $c$ online using a rule analogous to Eq.\ref{eq:z_learning_rule}, 
\begin{equation}\label{eq:learning_C}
    c(o_{t}) \leftarrow (1 - \eta_{t})~c(o_{t}) + \eta_{t}~exp(r_{t})~c(o_{t+1}).
\end{equation}
In the above equation, $c(o_{t})$ represents the desirability of an observation $o$ at time $t$. The value of $c(o_{t})$ is updated depending on the reward received and the desirability of the observation received at the next time step $c(o_{t+1})$.

The learning rate $\eta$ is a time-dependent parameter in Z-learning, as given in the equation below. $e$ is a hyperparameter we optimise that influences how fast/slow $\eta$ gets updated over time \cite{Todorov2009}:

\begin{equation}
\label{eq:learning_rule_c}
    \eta_{t} = \frac{e}{e + t}.
\end{equation}

If $\eta_t$ is high, the algorithm puts more weight on the new information. If $\eta_t$ is low, the algorithm puts more weight on the current estimate. Using the update rule in Eq.\ref{eq:learning_C} with the learning rate evolving as in \eqref{eq:learning_rule_c}, the value of $c$ evolves over time and may be used to update $\mathbb{C}$ online, ensuring that $\mathbb{C}$ is a categorical distribution over observations using the softmax function:
\begin{equation}
    \mathbb{C} = \sigma(c).
\end{equation}

\begin{figure}
    \centering
    \includegraphics[width = 0.8\textwidth]{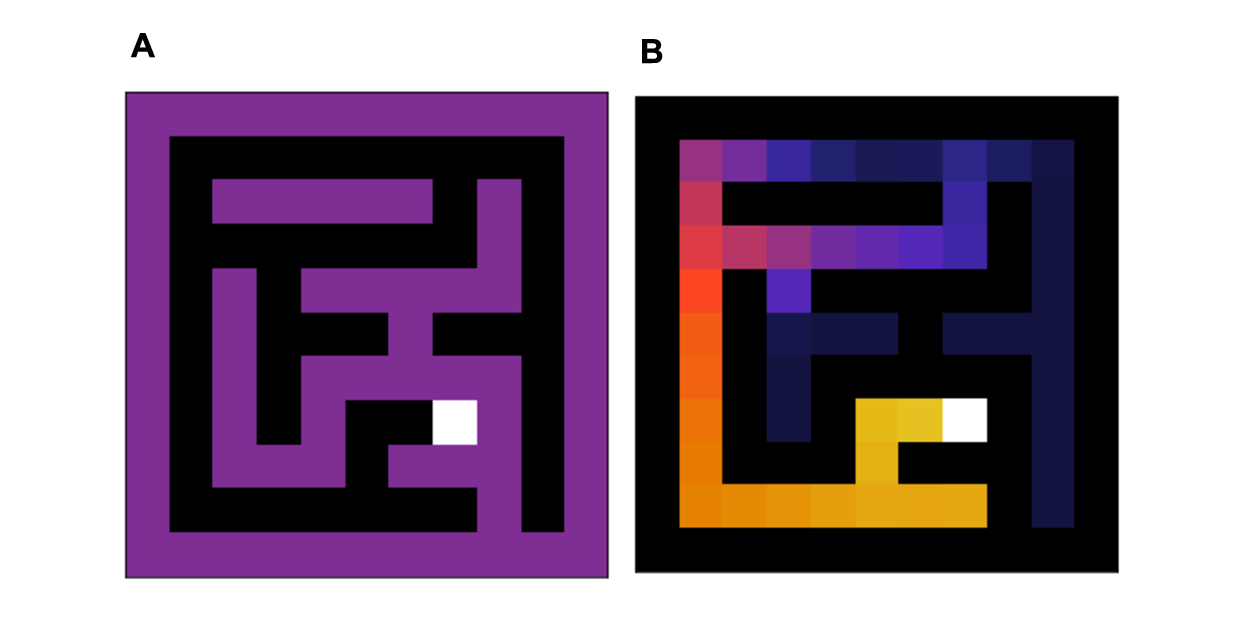}
    \caption{A: The sparsely defined preference distribution used by DPEFE agent in simulations, B: The learned preference distribution by AIF (\textbf{T}=1) agent over 50 episodes. Lighter colours imply a higher preference for the corresponding states.}
    \label{fig:learned_c}
\end{figure}

We use the standard grid world environments as shown in Fig.\ref{fig:grids} for the evaluation of the performance of various agents (more details in the next sections). Fig.\ref{fig:learned_c}, is a visualisation that represents the learned prior preference (for the grid shown in Fig. \ref{fig:grids} (A)) useful for the sophisticated inference agent. With an informed prior preference like this, the agent needs to plan only one time step ahead to navigate the grid successfully. It should be noted that in the DPEFE setting, we fix the prior preference $\mathbb{C}$ either before a trial or learn it when we encounter the goal as a one-hot vector. We are not learning an informed prior preference for the DPEFE agent in the simulations presented in the paper. The method for learning prior preference discussed in this section holds for any agent, but in our paper, DPEFE is not using this feature to demonstrate its ability to plan deeper. When we aid an active inference algorithm with the learning rule for $\mathbb{C}$, a planning horizon of $T = 1$ suffices to take desirable actions (i.e. with no deep tree search like in SI or policy space ($\Pi$) as in CAIF). With considering only the next time step, (i.e. only the consequence of immediately available actions), planning in all active inference agents (CAIF, SI, and DPEFE) are algorithmically equivalent. In the rest of the paper, we call this agent with planning horizon $T = 1$, which is aided with the learning rule of $C$ as active inference AIF (\textbf{T} = 1) agent. In our simulations, we compare the performance of these two approaches (i.e., deep planning with sparse $C$ and short-term planning with learning $C$).
An animation that visualises the learning prior preference distribution for the grid over $50$ episodes in Fig.\ref{fig:prior_saif} can be found in this \href{https://github.com/aswinpaul/dpefe_2023}{link}. In the following section, we discuss and compare the computational complexity of planning between existing and newly introduced schemes.

\section{Computational complexity}\label{compcomplex}

In this section, we compare the computational complexity in evaluating the expected free energy term, used for planning and decision making, with two other active inference approaches: classical active inference \cite{DaCosta2020},\cite{Sajid2021}, and sophisticated inference \cite{Friston2021}. 

In classical active inference (\cite{DaCosta2020}, \cite{Sajid2021}), the expected free energy for an MDP (i.e., a fully observable case) is given by, 
\begin{equation}\label{efecai}
    G(\pi \vert s_{t-1}) = D_{KL}[Q(s_{t} \vert \pi) \vert \vert P(s_{t})].
\end{equation}
Here, $P(s_t)$ represents an agent's prior preference and is equivalent to $\mathbb{C}$ in an MDP setting. In this paper, $\mathbb{C}$ is directly defined in terms of the hidden states. To avoid confusion, we always use the notation $\mathbb{C}$ in this paper regarding the observations $o$.

Similarly, for sophisticated inference \citep{Friston2021}, we have,
\begin{equation}\label{efesi}
    G(u_{t}) = D_{KL}[Q(s_{t+1} \vert u_{< t+1}) \vert \vert P(s_{t +1})] + \mathbb{E}_{Q(u_{t+1})}[G(u_{t+1})].
\end{equation}
In the above equation, we restrict the recursive evaluation of the second term, forward in time, till a `planning horizon (T)' as mentioned in \cite{Friston2021}. $T$ necessary for 'full-depth planning' i.e planning to the end of the episode is often required for sparsely defined prior preferences. This is required since the agent would not be able to differentiate the desirability of actions until reaching the last step of the episode through a tree search.

In classical active inference, to evaluate Eq.\ref{efecai}, the computational complexity is proportional to: 
$\mathcal{O} [\mathbf{card}(S) \times \mathbf{card}(U)^{T}].$    
For sophisticated inference, to evaluate Eq.\ref{efesi}, the complexity scales proportionally to: 
$    \mathcal{O} [(\mathbf{card}(S) \times \mathbf{card}(U)) ^ T].$ The dimensions of the quantities involved are specified in Tab.\ref{tab:dimension}. And recall that both Eq.\ref{efecai} and Eq.\ref{efesi} ignore the `ambiguity' term for simplicity.

\begin{table}[ht]
\centering
\begin{tabular}{|cc|cc|}
\hline
\multicolumn{2}{|c|}{Sophisticated inference}                                                                                        & \multicolumn{2}{|c|}{Classical active inference}                                                                \\ \hline
\multicolumn{1}{|c|}{Term}                                                           & Dimension                                       & \multicolumn{1}{c|}{Term}                            & Dimension                                               \\ \hline
\multicolumn{1}{|c|}{$s_{\tau+1}$}                                                     & $\mathbf{card}(S)$ (cardinality of S)                             & \multicolumn{1}{c|}{$\pi$}                           & $\mathbf{card}(U)^{T}$                                  \\ \hline
\multicolumn{1}{|c|}{$Q(u_{t+1})$}                                                & $\mathbf{card}(U)$                              & \multicolumn{1}{c|}{$s_{t}$}                      & $\mathbf{card}(S)$                                      \\ \hline
\multicolumn{1}{|c|}{$Q(s_{t+1} \vert u_{<t+1})$}                              & $\mathbf{card}(S)$                              & \multicolumn{1}{c|}{$Q(s_{t} \vert \pi)$}         & $\mathbf{card}(S) \times \mathbf{card}(U)^{T} $         \\ \hline
\multicolumn{1}{|c|}{$P(s_{\tau+1})$}                                                & $\mathbf{card}(S)$                              & \multicolumn{1}{c|}{$P(s_{t})$}                   & $\mathbf{card}(S)$                                      \\ \hline
\multicolumn{1}{|c|}{$G(u_{t+1})$}                                            & $\mathbf{card}(S)$                              & \multicolumn{1}{c|}{$G(\pi \vert t)$}             & $\mathbf{card}(U)^{T}$                                  \\ \hline

\multicolumn{2}{|c|}{$ \mathcal{O} [(\mathbf{card}(S) \times \mathbf{card}(U)) ^ T] $}                                                                                          & \multicolumn{2}{|c|}{$ \mathcal{O} [\mathbf{card}(S) \times \mathbf{card}(U)^{T}] $}                                                              \\ \hline

\end{tabular}
\caption{Computational complexity in evaluating EFE.}
\label{tab:dimension}
\end{table}

For evaluating EFE using dynamic programming, the expected free energy for an MDP can be deduced from Eq.\ref{eq.rec.pl} as,

\begin{equation} \label{efedpmdp}
G(u_{t} \vert s_{t}) = D_{KL}[Q(s_{t+1} \vert s_{t}, u_{t}) \vert \vert P(s_{t+1})] + \mathbb{E}_{Q(s_{t+1} \vert s_{t}, u_{t})}[G(u_{t+1} \vert s_{t+1})].
\end{equation}
Since we only evaluate on time-step ahead in Eq.\ref{efedpmdp}, even when evaluating backwards in time, the complexity scales as: $    \mathcal{O} [\mathbf{card}(S) \times \mathbf{card}(U) \times T].$

\section{Simulations results}
\label{sec:simulations}

\subsection{Setup}

\begin{figure}[t!]
    \centering
    \includegraphics[width = \textwidth]{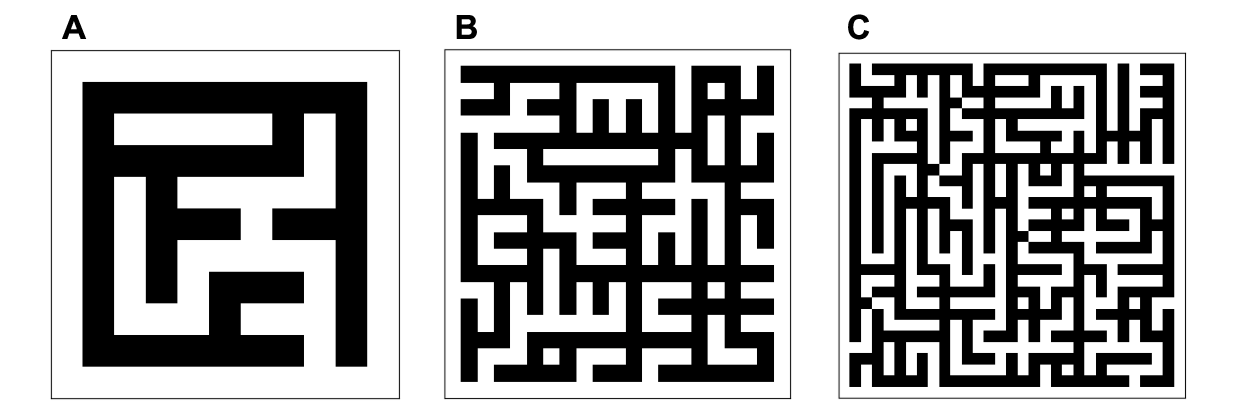}
    \caption{A: A standard grid world of 100 states with 50 valid states. B: A grid of 400 states with 204 valid states. C: A grid of 900 states with 497 valid states. These three grids are used for evaluating the performance of various schemes.}
    \label{fig:grids}
\end{figure}

We perform simulations in the standard grid world environment in Fig.\ref{fig:grids} to evaluate the performance of our proposed algorithms. The agent is born in a random start state at the beginning of every episode and can take one of the four available actions (North, South, East, West) at every time step to advance towards the goal state until the episode terminates either by a time-out ($10000$, $20000$, and $40000$ steps for the grids in Fig.\ref{fig:grids} respectively) or by reaching the goal state. For completeness, we compare the performance of the following algorithms in the grid world:
\begin{itemize}
    \item Q-learning: a benchmark model-free RL algorithm \cite{watkins1992q}
    \item Dyna-Q: a benchmark model-based RL algorithm improving upon Q-learning \cite{peng1993efficient}
    \item DPEFE algorithm with strictly defined (sparse) $\mathbb{C}$ (See Sec.\ref{dpefe_algorithm})
    \item Active inference algorithm aided with learning rule for $\mathbb{C}$ (See Sec.\ref{sec:learning_c}) and planning horizon of $T = 1$ i.e with no deep tree search like in SI, or policy space ($\Pi$) as in CAIF. With considering only the next time step, (i.e. only the consequence of immediately available actions) planning in all active inference agents (CAIF, SI, and DPEFE) are algorithmically equivalent. In the rest of the paper, we call this agent with planning horizon $T = 1$ aided with the learning rule of $C$ as active inference AIF (\textbf{T} = 1) agent.
\end{itemize}

We perform simulations in deterministic and stochastic grid variations shown in Fig.\ref{fig:grids}. The deterministic variation is a fully observable grid with no noise. So, an agent fully observes the present state---i.e., an MDP setting. Also, the outcomes of actions are non-probabilistic with no noise---i.e., a deterministic MDP setting. In the stochastic variation, we make the environment more challenging to navigate by adding $25\%$ noise in the transitions and $25\%$ noise in the observed state. In this case, the agent faces uncertainty at every time step about the underlying state (i.e., partially observable) and the next possible state (i.e., stochastic transitions)---i.e., a stochastic POMDP setting.

\subsection{Summary of results}

\begin{figure}[t!]
    \centering
    \includegraphics[width=\textwidth]{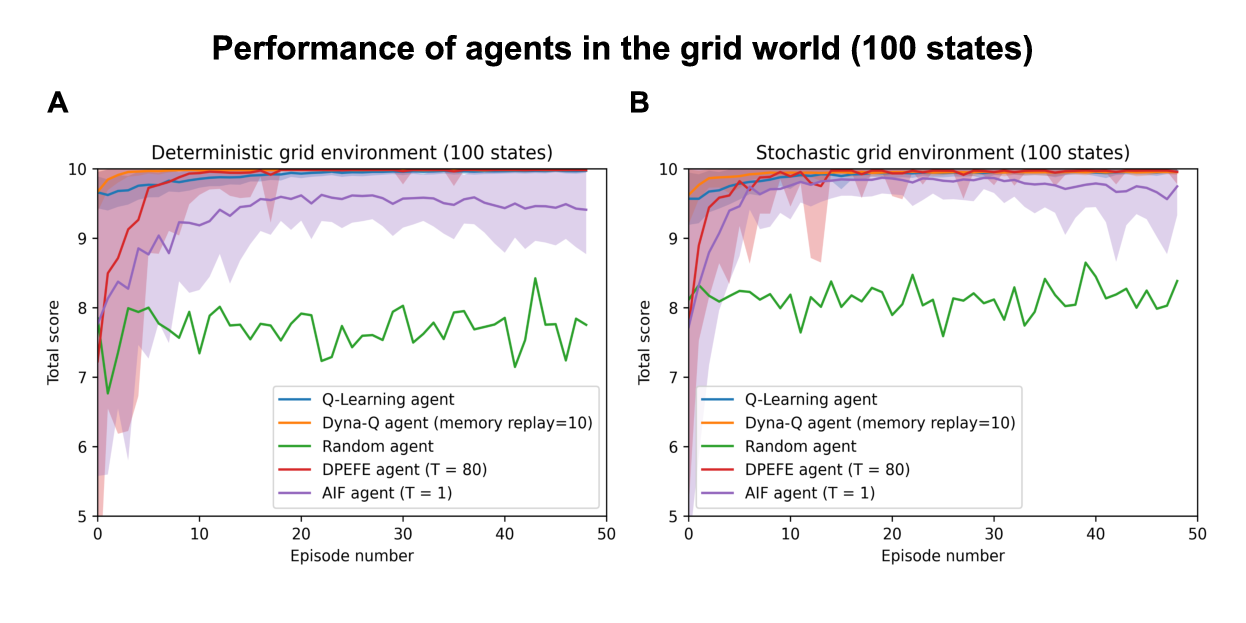}
    \caption{The summary of agents' performance in the two grids. A: Deterministic grid (100 states), B: Stochastic version of the grid in A (100 states, partially observable, stochastic transitions (POMDP)).}
    \label{fig:perf_agents_100}
\end{figure}

\begin{figure}[t!]
    \centering
    \includegraphics[width=\textwidth]{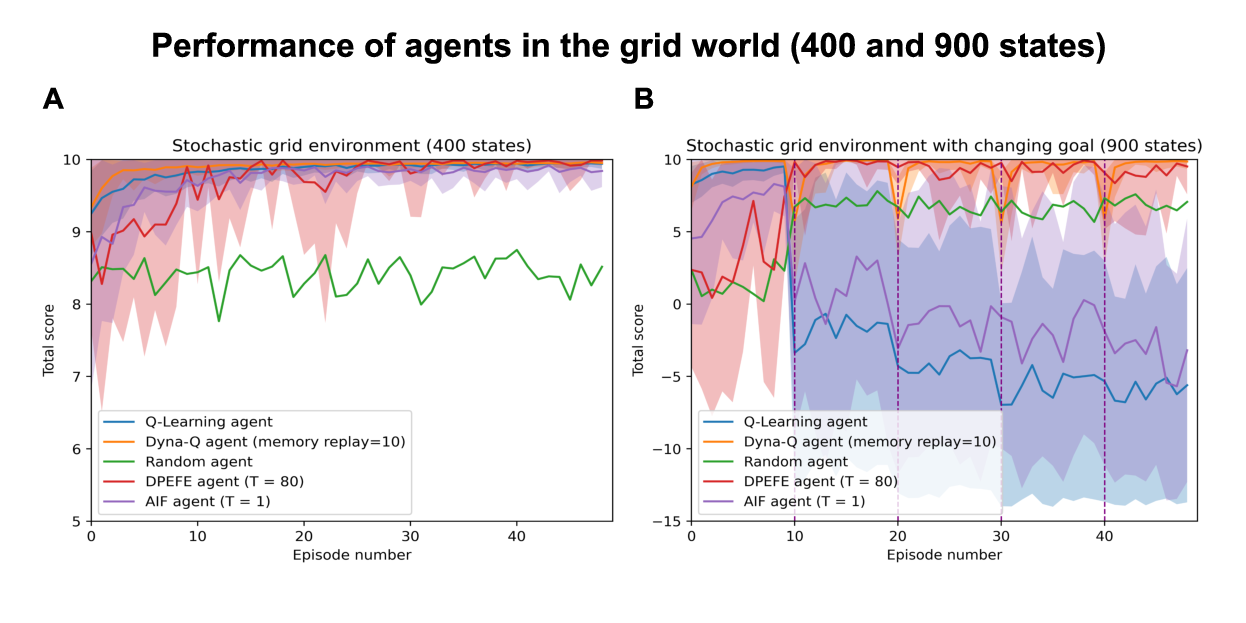}
    \caption{The summary of agents' performance in the two grids. A: Stochastic grid (400 states, partially observable, stochastic transitions (POMDP)), B: Stochastic grid in A with goal-state randomized after every ten episodes.}
    \label{fig:perf_agents}
\end{figure}

The agents' performance in this navigation problem is summarised in Fig.\ref{fig:perf_agents_100} and Fig.\ref {fig:perf_agents}. Performance is quantified in terms of how quickly an agent learns to solve the grid task, i.e. the total score. The agent receives a reward of ten points when the goal state is reached and a small negative reward for every step taken. The total score hence represents how fast the agent navigated to the goal state for a given episode. The grid has a fixed goal state throughout the episodes in Fig.\ref{fig:perf_agents_100} (A, B) and Fig.\ref{fig:perf_agents} (A). For simulations in Fig.\ref{fig:perf_agents} (B), the goal state is shifted to another random state every 10 episodes. This setup helps to evaluate the adaptability of agents in the face of changes in the environment. It is clear that during the initial episodes, the agents take longer to reach the goal state but learn to navigate quicker as the episodes unfold.
Standard RL algorithms (i.e., Dyna-Q and Q-Learning) are used here to benchmark the performance of active inference agents, as they are efficient state-of-the-art algorithms to solve this sort of task. 

In our simulations, the DPEFE algorithm performs at par with the Dyna-Q algorithm with a planning depth of $T = 80$ \footnote{a planning horizon more than any optimal path in this grid. Since the start state is randomized, optimal paths can have many lengths in the grid. A planning depth of $T=80$ ensures that the agent plans enough not to miss the length that needs to be covered in any setting.} (See (Fig.\ref{fig:perf_agents_100} (A, B), Fig.\ref{fig:perf_agents} (A)). The DPEFE agent performs even better when we randomised goal-states every $10$ episode (Fig.\ref{fig:perf_agents} (B)). In contrast to online learning algorithms like Dyna-Q, active inference agents can take advantage of the re-definable explicit prior preference distribution $\mathbb{C}$. For the AIF(\textbf{T = 1}) agent, we observe that the performance improves over time but is not as good as the DPEFE agent. This is because the AIF(\textbf{T = 1}) agent plans for only one step ahead in our trials by design. We could also observe that the Q-Learning agent performs worse than the random agent and recovers slower than the AIF(\textbf{T = 1}) agent when faced with uncertainty in the goal state. It is a promising direction to optimise the learning of prior preference $\mathbb{C}$ in the AIF(\textbf{T = 1}) agent, ensuring accuracy in the face of uncertainty. All simulations were performed for $100$ trials with different random seeds to ensure the reproducibility of results.

Besides this, we observe a longer time to achieve the goal state for both active inference agents (even longer than the `Random agent') in the initial episodes. This is a characteristic feature of active inference agents, as their exploratory behaviour dominates during the initial trials. The goal-directed behaviour dominates only after the agent sufficiently minimises uncertainty in the model parameters \cite{Tschantz2020}.

\subsection{Optimising learning rate parameter for AIF (\textbf{T} = 1) agent}

The learning rule proposed for sophisticated inference in Eq.\ref{eq:learning_rule_c} requires a (manually) optimised value of $e$ for every environment that influences the learning rate $\eta$. \cite{Todorov2009} inspires this learning rule, where the value of $\eta_t$ determines how fast the parameter $c$ converges for a given trial. The structure of learned $c$ is crucial for the active inference agent, as $\mathbb{C}$ determines how meaningful the planning is for the agent. In Fig.\ref{fig:eta_opt}, we plot the performance of the AIF(\textbf{T} = 1) agent as a function of $e$ for the grids in Fig. \ref{fig:grids}. A promising direction for future research is to improve the learning rule based on $\eta$ and fine-tune the method for learning $\mathbb{C}$. The observation in Fig.\ref{fig:eta_opt} is that the performance of the AIF(\textbf{T} = 1) agent is not heavily dependent on the value of $e$. We used different values of $e > 10000$ in AIF(\textbf{T} = 1) agents in all settings in this paper. 

\subsection{An emphasis on computational complexity}

\begin{figure}[t!]
    \centering
    \includegraphics[width=\textwidth]{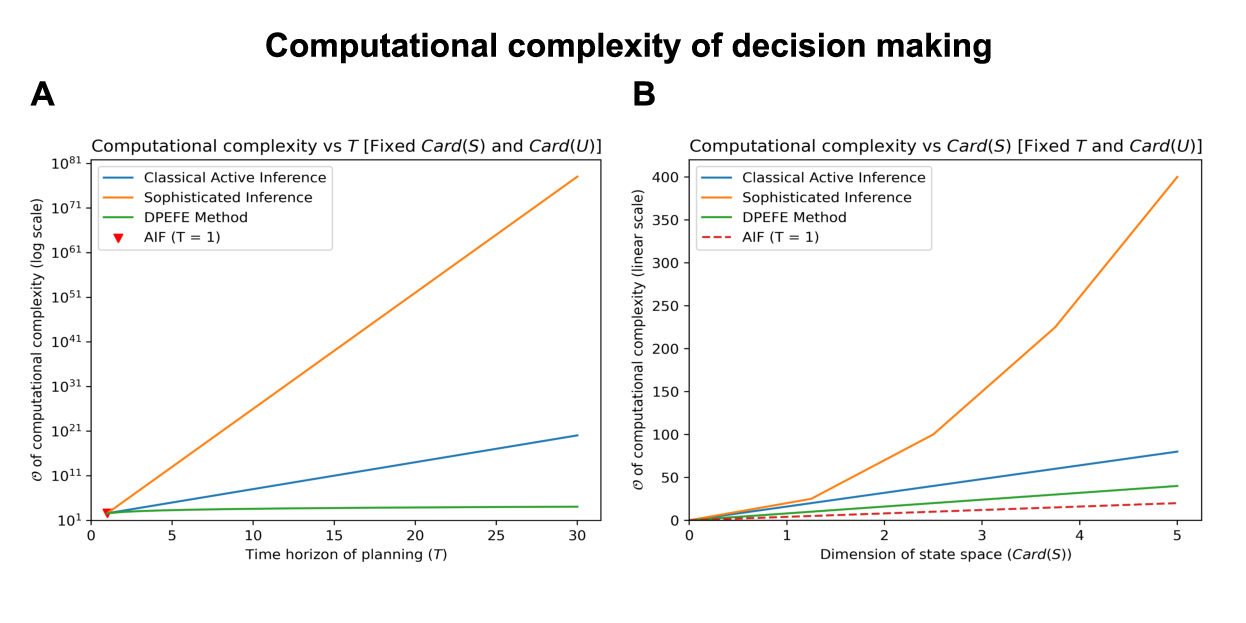}
    \caption{Comparing computational complexity of methods discussed in this paper. A: Order of computational complexity (log scale) vs Time horizon of planning (T). Here $\mathbf{card}(S) = 100$, $\mathbf{card}(U) = 4$., B: Order of computational complexity (linear scale) vs $\mathbf{card}(S)$. Here, $\mathbf{card}(U) = 4$ and $T = 2$ (T=1 for AIF (T=1) agent). We observe that except for DPEFE and AIF (T=1) methods, computational complexity becomes intractable even for $T$ as small $T=2$ and $\mathbf{card}(S)=5$.}
    \label{fig:c_complexity}
\end{figure}

To understand why the classical active inference (CAIF) and SI methods cannot solve these grid environments with the traditional planning method, we provide an exemplar setting in Tab. \ref{tab:egcompcomplex}. Consider the small grid as shown in  Fig.\ref{fig:grids} with $\mathbf{card}(S) = 100$, $\mathbf{card}(U) = 4$, $\textbf{T}=30$. Tab.\ref{tab:egcompcomplex} summarises the computational complexity of simulating various active inference agents for this small grid world problem. The computational complexity exceeds practical implementations even with a $\textbf{T} = 2$ planning horizon. We can observe this visually in Fig.\ref{fig:c_complexity}.

\begin{table}[ht]
\centering
\begin{tabular}{|c|c|c|}
\hline
Method                                                                                               & Order dimension ($\mathcal{O}$)                                                        & Approx. $\mathcal{O}$ for specific case \\ \hline
CAIF (T = 30) & $ \mathcal{O} [\mathbf{card}(S) \times \mathbf{card}(U)^{T}] $ & $10^{18}$            \\ \hline
SI (T = 30)                                           & $ \mathcal{O} [(\mathbf{card}(S) \times \mathbf{card}(U)) ^ T] $                       & $10^{68}$             \\ \hline
$\text{SI (T = 2)}$                                                                          & $ \mathcal{O} [\mathbf{card}(S) \times \mathbf{card}(U) ^ 2]. $                  & $10^6$              \\ \hline
$\text{DPEFE (T = 30)}^{*}$                                                                           & $ \mathcal{O} [\mathbf{card}(S) \times \mathbf{card}(U) \times T]. $                  & $10^3$              \\ \hline
$\text{AIF (\textbf{T} = 1)}^{*}$                                                                          & $ \mathcal{O} [\mathbf{card}(S) \times \mathbf{card}(U)]. $                  & $0.2 * 10^3$              \\ \hline
\end{tabular}
\caption{Computational complexity for evaluating EFE with $\mathbf{card}(S) = 100$, $\mathbf{card}(U) = 4$, $T=30$. CAIF: classical active inference, SI: sophisticated inference with full tree search, DPEFE: dynamic programming active inference agent, AIF (\textbf{T} = 1): Active inference agent planning one time-step ahead and learning the prior preference, *proposed in this paper. Without tree search, the sophisticated inference (SI) agent is algorithmically equivalent to the classical active inference agent (CAIF). In the rest of the paper, we call this agent with planning horizon $T = 1$ as active inference AIF(\textbf{T} = 1) agent.}
\label{tab:egcompcomplex}
\end{table}

However, we note that the proposed solution of first learning the prior preferences (See Sec.\ref{sec:learning_c}) using the Z-learning rule enables the active inference (AIF) agent to learn and solve the unknown environment by avoiding the computational complexity of a deep tree search. It should also be noted that neither of the active inference algorithms (DPEFE and AIF (\textbf{T}=1)) was equipped with meaningful priors about the (generative) model parameters ($\mathbb{B}$, $\mathbb{C}$, and $\mathbb{D}$). Agents start blindly with `uninformed' model priors and evolve by integrating all aspects of behaviour: perception, planning, decision-making, and learning. Yet, the fact that, like Dyna-Q, they start with a model of the world means that they are much less agnostic than the model-free alternative offered by Q-learning. The following section discusses the merits and limitations of the proposed solutions to optimise decision-making in active inference.

\section{Discussion}

In this work, we explored the usefulness of active inference as an algorithm to model intelligent behaviour and its application to a benchmark control problem, the stochastic grid world task. We identified the limitations of some of the most common formulations of active inference \cite{Friston2021}, which do not scale well for planning and decision-making tasks in high-dimensional settings. We proposed two computational solutions to optimise planning: harnessing the machinery offered by dynamic programming and the Bellman optimality principle and harnessing the Z-learning algorithm to learn informed preferences. 

First, our proposed planning algorithm evaluates expected free energy backwards in time, exploiting Bellman's optimality principle, considering only the immediate future as in the dynamic programming algorithm. We present an algorithm for general sequential POMDP problems that combines perception, action selection and learning under the single cost function of variational free energy. Additionally, the prior preference, i.e., the goal state about the control task, was strictly defined (i.e., uninformed) and supplied to the agent, unlike well-informed prior preferences as seen in earlier formulations.
Secondly, we explored the utility of equipping agents so as to learn their prior preferences. We observed that learning the prior preference enables the agent to solve the task while avoiding the computationally (often prohibitively) expensive tree search. We used state-of-the-art model-based reinforcement learning algorithms, such as Dyna-Q, to benchmark the performance of active inference agents.
Lastly, there is further potential to optimise computational time by exploiting approximation parameters involved in planning and decision-making. For example, the softmax functions used while planning and decision-making determine the precision of output distributions. There is also scope to optimise further the SI agent proposed in this paper by learning the prior preference. Based on the Z-learning method, the learning rule for prior preference parameters shall be optimised and fine-tuned for active inference applications in future work. Since the Z-learning method is fine-tuned for a particular class of MDP problems \cite{Todorov2006}, we leave a detailed comparison of the two approaches to future work. We conclude that the above results advance active inference as a promising suite of methods for modelling intelligent behaviour and for solving stochastic control problems.

\section{Acknowledgments}
AP acknowledges research sponsorship from IITB-Monash Research Academy, Mumbai and the Department of Biotechnology, Government of India. AR is funded by the Australian Research Council (Refs: DE170100128 \& DP200100757) and Australian National Health and Medical Research Council Investigator Grant (Ref: 1194910). AR is a CIFAR Azrieli Global Scholar in the Brain, Mind \& Consciousness Program. AR, NS, and LD are affiliated with The Wellcome Centre for Human Neuroimaging, supported by core funding from Wellcome [203147/Z/16/Z]. NS is funded by the Medical Research Council (MR/S502522/1) and the 2021-2022 Microsoft PhD Fellowship. LD is supported by the Fonds National de la Recherche, Luxembourg (Project code: 13568875). This publication is based on work partially supported by the EPSRC Centre for Doctoral Training in Mathematics of Random Systems: Analysis, Modelling and Simulation (EP/S023925/1).

\section{Software note}

All the code for agents, optimisation and grid environments used are custom written in Python 3.9.15 and is available in this project repository: \url{https://github.com/aswinpaul/dpefe_2023}.

\bibliographystyle{unsrtnat}
\bibliography{bib_dpefe}

\newpage
\appendix
\numberwithin{equation}{section}
\numberwithin{figure}{section}

\section{Derivation of optimal state-belief}
\label{appendix:perception_derivation}
We want to differentiate the following w.r.t. $Q(s)$:

\begin{equation}
    F = \sum_{s} Q(s)[~\text{log} Q(s)-\text{log} P(o\vert s) - \text{log}P(s)~].
\end{equation}
First, note that the derivative of the logarithm function is $\frac{1}{x}$. Second, observe that the derivative of $Q(s)$ with respect to $Q(s)$ is 1. With these two pieces in mind, we can differentiate $F$:

Let's define $f(s) = \log Q(s) - \log P(o|s) - \log P(s)$, then:

\begin{equation}
\frac{dF}{dQ(s)} = \sum_{s} [~\frac{df}{dQ(s)} \cdot Q(s) + f(s) \cdot \frac{dQ(s)}{dQ(s)}~]
\end{equation}

The derivative of $f(s)$ with respect to $Q(s)$ can be computed as:

\begin{equation}
\frac{df}{dQ(s)} = \frac{1}{Q(s)} - 0 - 0 = \frac{1}{Q(s)}
\end{equation}

This leads to:

\begin{align}
\frac{dF}{dQ(s)} &= \sum_{s} [~Q(s) \cdot \left(\frac{1}{Q(s)}\right) + f(s)~] \\
&= \sum_{s} 1 + \log Q(s) - \log P(o|s) - \log P(s)
\end{align}

So, the derivative of $F$ with respect to $Q(s)$ is:

\begin{equation}
\frac{dF}{dQ(s)} = \sum_{s} 1 + \log Q(s) - \log P(o|s) - \log P(s)
\end{equation}

The goal is to minimize the free energy $F$ with respect to the distribution $Q(s)$. To find the minimum, we can set the derivative of $F$ with respect to $Q(s)$ to zero. From the previous derivation, we know that:

\begin{equation}
\frac{dF}{dQ(s)} = \sum_{s} 1 + \log Q(s) - \log P(o|s) - \log P(s)
\end{equation}

Setting this equal to zero gives:

\begin{align}
1 + \log Q(s) - \log P(o|s) - \log P(s) = 0 \\
\log Q(s) = \log P(o|s) + \log P(s) - 1
\end{align}

However, note that the log function is typically normalized such that the sum of the probabilities in $Q(s)$ equals $1$ (since $Q(s)$ is a probability distribution), so we can safely ignore the $-1$ term:

\begin{equation}
\log Q(s) = \log P(o|s) + \log P(s)
\end{equation}

The optimal distribution $Q^*(s)$ that minimizes the free energy $F$ is thus:

\begin{equation}
\log Q^*(s) = \log P(o|s) + \log P(s)~.
\end{equation}

\newpage
\section{Optimising learning parameter for AIF (T=1) agent}

\begin{figure}[hbt!]
    \centering
    \includegraphics[width = \textwidth]{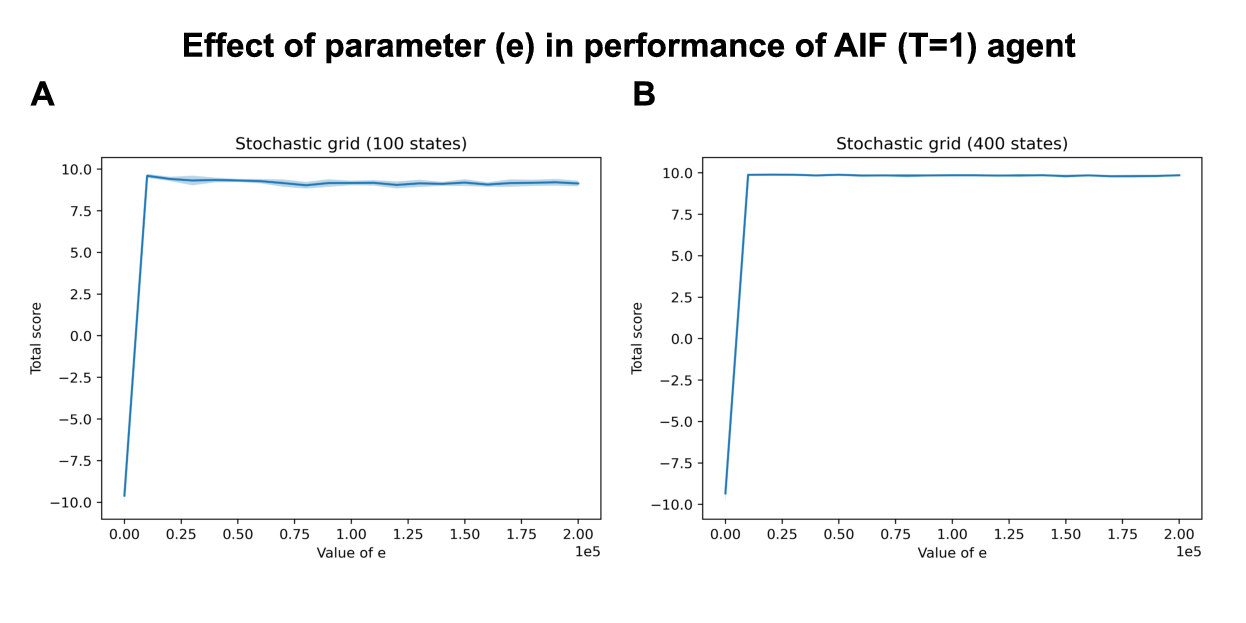}
    \caption{A sample graph of manual optimisation for $e$ for Z-learning in the AIF(T=1) algorithm. A: For a stochastic grid with 100 states, B: For a stochastic grid with 400 states. We observe that the agent's performance is not heavily dependent on the value of $e$ that controls the learning parameter $\eta_{t}$.} 
    \label{fig:eta_opt}
\end{figure}





\end{document}